\documentclass[10pt,twocolumn,letterpaper]{article}

\usepackage[pagenumbers]{wacv} 

\usepackage{graphicx}
\usepackage{amsmath}
\usepackage{amssymb}
\usepackage{booktabs}

\usepackage[pagebackref,breaklinks,colorlinks]{hyperref}

\usepackage{enumitem}
\usepackage{booktabs}
\usepackage{graphicx}
\usepackage{algorithm}
\usepackage[noend]{algpseudocode}

\usepackage{bm}
\usepackage{xspace}

\newcommand{\cls}[1]{{\small\texttt{#1}}}
\newcommand{\ours}{{\fontfamily{cmr}\selectfont EMILIE}\xspace}

\newcommand{\ourbenchmark}{IMIE-Bench\xspace}

\usepackage[capitalize]{cleveref}
\crefname{section}{Sec.}{Secs.}
\Crefname{section}{Section}{Sections}
\Crefname{table}{Table}{Tables}
\crefname{table}{Tab.}{Tabs.}

\def\wacvPaperID{2458} 
\def\confName{WACV}
\def\confYear{2024}

\begin{document}

\title{Iterative Multi-granular Image Editing using Diffusion Models}
\author{K J Joseph, Prateksha Udhayanan, Tripti Shukla, Aishwarya Agarwal, Srikrishna Karanam, \\ Koustava Goswami, Balaji Vasan Srinivasan\\
Adobe Research\\
{\tt\small \{josephkj,udhayana,trshukla,aishagar,skaranam,koustavag,balsrini\}@adobe.com}
}

\twocolumn[{
\renewcommand\twocolumn[1][]{#1}%
\maketitle
\begin{center}
 \centering
 \captionsetup{type=figure}
 \includegraphics[width=1.0\textwidth]{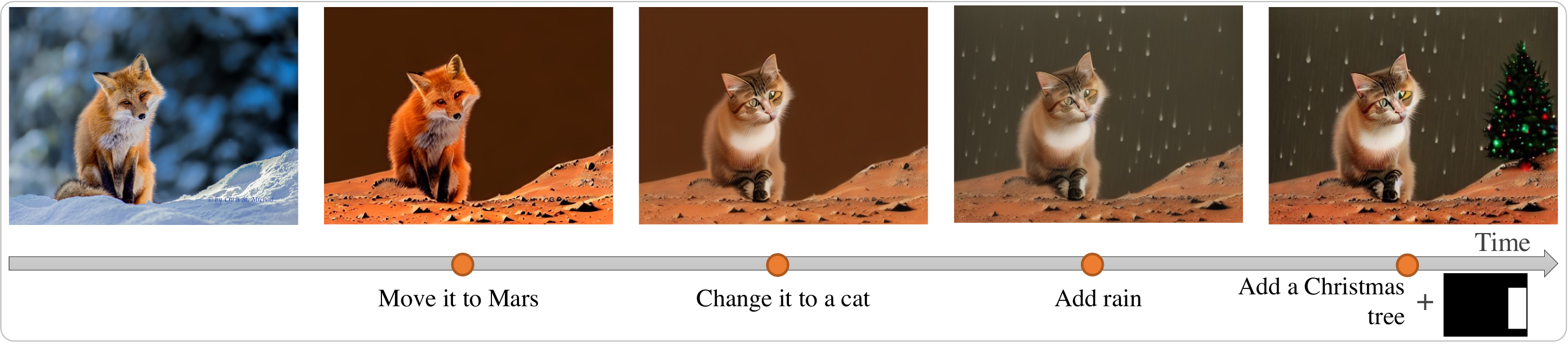}
 \caption{We introduce \ours: It\underline{e}rative \underline{M}ult\underline{i} granu\underline{l}ar \underline{I}mage \underline{E}ditor, a diffusion model that can faithfully follow a series of image editing instructions from a user. In this example, we see how the image of the \texttt{fox} has been semantically modified according to the provided edit instructions. Optionally, a user can 
 instruct \ours with exact location of the desired edit, as illustrated in the last column.}
 \label{fig:teaser}
\end{center}
}]

\maketitle

\begin{abstract}
    \vspace{-5pt}
  Recent advances in text-guided image synthesis has dramatically changed how creative professionals generate artistic and aesthetically pleasing visual assets. 
  To fully support such creative endeavors, the process should possess the ability to: 1) iteratively edit the generations and 2) control the spatial reach of desired changes (global, local or anything in between). We formalize this pragmatic problem setting as \cls{Iterative Multi-granular Editing}.
  While there has been substantial progress with diffusion-based models for image synthesis and editing, they are all one shot (i.e., no iterative editing capabilities) and do not naturally yield multi-granular control (i.e., covering the full spectrum of local-to-global edits). 
  To overcome these drawbacks, we propose \textit{\cls{\ours}: It\underline{e}rative \underline{M}ult\underline{i}-granu\underline{l}ar \underline{I}mage \underline{E}ditor}.
  \ours introduces a novel latent iteration strategy, which re-purposes a pre-trained diffusion model to facilitate iterative editing. This is complemented by a gradient control operation for multi-granular control.
  We introduce a new benchmark dataset to evaluate our newly proposed setting. We conduct exhaustive quantitatively and qualitatively evaluation against recent state-of-the-art approaches adapted to our task, to being out the mettle of \ours. We hope our work would attract attention to this newly identified, pragmatic problem setting.
  
\end{abstract}


\section{Introduction}\label{sec:introduction}
An image is an invaluable form of visual communication. Creating such a visual illustration is a creative process and an expression of the ingenuity of the artist. They often start with a blank canvas and \textit{iteratively} update it with the semantic concepts that they want to convey. These changes might be at different \textit{granularities}, ranging from any small local change to global changes spanning the entire canvas.

Generative technologies for image synthesis have made remarkable strides lately. Diffusion models \cite{song2020denoising,ho2020denoising,rombach2022high} have had tremendous success in creating realistic images from text prompts. These text-to-image models have the capacity to inspire and augment human creativity.
Diffusion models for image editing \cite{tumanyan2022plug,mokady2022null,kawar2022imagic,couairon2023diffedit,avrahami2022blended_latent,wang2022imagen,brooks2022instructpix2pix} further enhances the collaborative content creation, by allowing users to edit their content using the versatility of diffusion models. 

In this work, we identify two major gaps in utilizing diffusion based image editing models as an assistant in a creator's workflow: $1$) Current diffusion based image-editing methods are one-shot. They consume an input image, make the suggested edit, and give back the result. This is in stark contrast to the creator's workflow, which is naturally iterative. 
$2$) It is not easy for an artist to specify and constrain the spatial extent of the intended edit. We note that this is a very practical yet under-explored research direction in the literature. 
Towards this end, we formalize and define this novel problem setting as \textit{Iterative Multi-granular Image Editing}.


A naive approach to iteratively edit an image would be to use the edited image from the previous step as input to the image editor for the next step. This, however, adds unwanted artifacts to the outputs, and these get accumulated over edit steps as shown in \cref{fig:motivating_latent_iteration,fig:comparison}. To address this, we introduce a new and elegant \textit{latent iteration} framework. Our key insight is that iterating over the latent space instead of the image space, there is a substantial reduction in the amount of noise/artifacts that get added over edit steps. Next, to incorporate multi-granular control, we  modulate the denoising process to follow the user-specified location constraints. We interpret the diffusion model as an energy-based model, which allows us to control the spatial extent of the edits by selectively restricting the gradient updates to the regions of interest. These two contributions, put together as part of our overall framework called \textit{\ours: It\underline{e}rative \underline{M}ult\underline{i}-granu\underline{l}ar \underline{I}mage \underline{E}ditor}, tackles our newly introduced problem setting. It is critical to note that \ours adds these capabilities directly into an already trained diffusion model and does not require any retraining, thereby substantially enhancing its usability.

As we propose and address a novel problem setting, we find that there are no existing benchmark datasets that we can use to evaluate the methodologies. Towards this end, we introduce a new benchmark dataset, called \ourbenchmark, particularly suited to our problem. We extensively evaluate \ours both qualitative and quantitatively on our proposed benchmark and show how it performs against various baselines adapted to this new problem. Further, we also evaluate the multi-granular editing capabilities of EMILIE on the publicly available EditBench \cite{wang2022imagen} benchmark. We see that our approach outperforms competing state-of-the-art methods like Blended Latent Diffusion \cite{avrahami2022blended_latent} and DiffEdit \cite{couairon2023diffedit} both qualitatively and quantitatively.


\noindent To summarize, the key highlights of our work are:

\begin{itemize}[leftmargin=*,topsep=0pt, noitemsep]
\item We introduce \textit{a novel problem setting} of iterative, multi-granular image editing motivated by practical, real-world use-cases of multimedia content designers.
\item To tackle our new problem, we propose \ours, a training-free approach that comprises a \textit{new latent iteration method} to reduce artifacts over iterative edit steps and a gradient update mechanism that enables controlling the spatial extent of the edits.
\item We propose a \textit{new benchmark dataset}, \ourbenchmark, particularly suited to our new problem setting comprising a carefully curated set of images with a sequence of at least four edit instructions each. 
\item We conduct \textit{exhaustive experimental evaluation} on \ourbenchmark and EditBench (for multi-granular edits) to bring out the efficacy of our proposed approach, where we clearly outperform the existing state-of-the-art methods.
\end{itemize}


\section{Related Works}\label{sec:related_works}


\noindent\textbf{Multimodal Image Generation and Editing Methods}: 
Generative Adversarial Networks (GAN) \cite{goodfellow2020generative} based text to images approaches like StackGAN \cite{zhang2017stackgan}, StackGAN++ \cite{zhang2018photographic} and AttnGAN \cite{xu2018attngan} were effective in generating $64 \times 64 \times 3$ images conditioned on their textual description. These methods work well in modelling simple objects like generating flowers and birds, but struggle to generate complex scenes with multiple objects. Recently, diffusion based methods \cite{dhariwal2021diffusion,rombach2022high} have had phenomenal success in generating realistic images from textual descriptions.

Similar to the success in image synthesis domain, diffusion based models have had significant strides for editing images too. Imagic \cite{kawar2022imagic} is able to make complex non-rigid edits to real images by learning to align a text embedding with the input image and the target text. PnP \cite{tumanyan2022plug}, NTI \cite{mokady2022null}, Imagic \cite{kawar2022imagic}, DiffEdit \cite{couairon2023diffedit} first use DDIM inversion \cite{dhariwal2021diffusion,song2020denoising} to invert the image to the input tensor required by the the diffusion model, and then use the text conditioned denoising process to generate the image with the required edit. CycleDiffusion \cite{wu2022unifying} proposes DPM-Encoder to project an image to the latent space of diffusion models thereby enabling them to be used as zero-shot image editors.
InstructPix2Pix \cite{brooks2022instructpix2pix} and Imagen Editor \cite{wang2022imagen} pass the image to be edited directly to the diffusion model by bypassing the DDIM inversion step. This improves the fidelity of the generated edits. But, none of these works addresses the challenges involved with \textit{iterative editing}, which is the main focus of our work. A trivial way to make these efforts iterative would be to pass the edited image recursively. We compare with such iterative variants of PnP \cite{tumanyan2022plug} and InstructPix2Pix \cite{brooks2022instructpix2pix} in \cref{sec:experiments}.

Recent efforts in mask guided image editing methods \cite{couairon2023diffedit,avrahami2022blended_latent} indeed can support multi-granular editing. We compare with these methods in \cref{sec:experiments}.

\vspace{5pt}\noindent\textbf{Iterative Image Synthesis}: This area is relatively less explored. CAISE \cite{kim2022caise} and Conversational Editing \cite{manuvinakurike2018conversational} applies operations like `rotate image by $90^\circ$', `change the contrast by $60$', `crop the image' and so on. SSCR \cite{fu2020sscr} and Keep drawing it \cite{el2018keep} operates on synthetic data from CLEVR \cite{johnson2017clevr} dataset, and tries to add objects to a canvas incrementally. Our approach operates on natural images, and can handle real-world edits that an artist might make to an image.

\section{Anatomy of Latent Diffusion Models} \label{sec:anatomy}
Diffusion models \cite{sohl2015deep,ho2020denoising} are a class of probabilistic models that generates a sample from a data distribution,  
$\bm{x_0} \sim p(\bm{x_0})$
by gradually denoising a normally distributed random variable $\bm{x}_T \sim \mathcal{N}(\bm{0}, \textbf{I})$, over $T$ iterations. The denoising process creates a series of intermediate samples $\bm{x}_T, \bm{x}_{T-1}, \cdots, \bm{x}_0$; with decreasing amount of noise. The amount of noise to be removed at each step is predicted by a neural network $\epsilon_\theta(\bm{x}_t, t)$, where $t$ is the denoising step.

These models has been found to be extremely successful in synthesizing realistic images \cite{dhariwal2021diffusion,saharia2022photorealistic,ramesh2022hierarchical} when $\bm{x}_i \in \mathbb{R}^{W \times H \times 3}$. To reduce the computational requirements for training without degrading quality and flexibility, Rombach \etal proposed Latent Diffusion \cite{rombach2022high}, which does the diffusion process in the latent space of a pretrained VQ-VAE \cite{van2017neural} encoder. Hence, the input image $\bm{x}_0$ is first projected into the latent space $\bm{z}_0 = \mathcal{E}(\bm{x}_0)$, where $\mathcal{E}(\cdot)$ is the VQ-VAE encoder and $\bm{z}_0 \in \mathbb{R}^{w \times h \times 4}$, where $w = W / 2^4$ and $h = H / 2^4$ \cite{rombach2022high}. 
The forward diffusion process uses a Markovian noise process $q$, which gradually adds noise to $\bm{z}_{0}$ through $\bm{z}_{T}$ with a Gaussian function following a variance schedule $\beta_t$:
\begin{equation}
    q(\bm{z}_{t} | \bm{z}_{t-1}) = \mathcal{N}(z_t; \sqrt{1 - \beta_t}\bm{z}_{t-1}, \beta_t \textbf{I})
\end{equation}
Ho \etal \cite{ho2020denoising} showed that sampling $\bm{z}_t \sim q(\bm{z}_{t} | \bm{z}_{0})$ need not be iterative, but instead can be as follows: 
\begin{align} 
q(\bm{z}_{t} | \bm{z}_{0}) &=  \mathcal{N}(z_t; \sqrt{\bar{\alpha}_t}\bm{z}_{0}, (1 - \bar{\alpha}_t) \textbf{I}) \\ 
 &=  \sqrt{\bar{\alpha}_t}\bm{z}_{0} + \epsilon \sqrt{(1 - \bar{\alpha}_t)}, \epsilon \sim \mathcal{N}(\boldsymbol{0}, \textbf{I}) \label{eqn:noise}
\end{align}
where $\alpha_t = 1 - \beta_t$ and $\bar{\alpha}_t = \prod_{r=0}^t \alpha_r$. The reverse diffusion process learns $\epsilon_\theta(\bm{z}_t, t)$ to predict $\epsilon$ from \cref{eqn:noise}. The learning objective is as follows:
\begin{equation}
    \mathcal{L}_{LDM} = \mathbb{E}_{t\sim[1,T], \bm{z}_0 = \mathcal{E}(\bm{x}_0), \epsilon \sim \mathcal{N}(\boldsymbol{0}, \textbf{I})}[\lVert \epsilon - \epsilon_\theta(\bm{z}_t, t) \rVert^2]
\end{equation}

To sample an image from a learned noise prediction function 
(loosely referred to as diffusion model henceforth) from $\bm{z}_T \sim \mathcal{N}(\boldsymbol{0}, \textbf{I})$, we iteratively apply the following step:
\begin{equation}
    \bm{z}_{t-1} = \bm{z}_{t} - \epsilon_\theta(\bm{z}_t, t) + \mathcal{N}(\bm{0}, \sigma_t^2\textbf{I}) \label{eqn:inference}
\end{equation}
where $\sigma_t = \sqrt{\beta_t}$. We refer the readers to Ho \etal \cite{ho2020denoising} for more details on the sampling process. Finally, $\bm{z}_0$ is projected back to the image space by the VQ-VAE decoder $\mathcal{D}$: $\bm{x}_0 = \mathcal{D}(\bm{z}_0)$.

The diffusion model $\epsilon_\theta(\cdot, \cdot)$ is generally implemented with a time-conditional UNet \cite{ronneberger2015u} architecture. It contains an encoder, middle layer and a decoder. Each of these modules contains multiple blocks with a residual convolutional layer, self-attention layer and cross-attention layer.
The diffusion model can be optionally augmented to consume an extra condition $\bm{y}$ (like text, caption, canny-maps, segmentation masks, skeletal information \etc) as follows: $\epsilon_\theta(\bm{z_t}, t, \bm{y})$. The cross-attention layers effectively warps the extra conditioning into the diffusion model. 

Latent Diffusion Models (LDM) have been effectively applied in image editing applications too. The major challenge is to make modifications specified via a textual condition $\bm{y}$ to an existing image $\bm{x}_0$. Methods like PnP \cite{tumanyan2022plug}, NTI \cite{mokady2022null}, Imagic \cite{kawar2022imagic}, DiffEdit \cite{couairon2023diffedit} first invert $\bm{x}_0$ to $\bm{z}_{inv}$ using techniques similar to DDIM Inversion \cite{dhariwal2021diffusion,song2020denoising}, so that when they start the diffusion process (in \cref{eqn:inference}) from $\bm{z}_{inv}$, they will be able to get back $\bm{x}_0$. Alternatively, approaches like InstructPix2Pix \cite{brooks2022instructpix2pix} and Imagen Editor \cite{wang2022imagen} learn few extra layers to directly pass $\bm{x}_0$ as input to the diffusion model. This approach retains better characteristics of the input image, as it side-steps the inversion process. We build \ours by extending InstructPix2Pix \cite{brooks2022instructpix2pix} to support iterative and multi-granular control, as it has minimal changes from LDM \cite{rombach2022high} (the input tensor $\bm{z}_t$ is augmented with four more channels to consume $\bm{x}_0$, and the corresponding weights are initialized to zero. No other changes to LDM \cite{rombach2022high}), and that its trained model is publicly available.

\section{Iterative Multi-granular Editing}
As previously noted in Section~\ref{sec:introduction}, existing techniques are all focused on one-shot generation (no iterative capabilities) and do not provide the ability for users to specify and constrain the spatial extent of the intended edits. Since creative professionals would want to iteratively edit images on the canvas while needing more spatial control on where in the image (global or local) the edits go, the first contribution of this work is the formulation of a new problem setting we call \textit{Iterative Multi-granular Image Editor}. Given an input image $\bm{I}_0$, a set of edit instructions $E = \{\bm{y}_1, \cdots, \bm{y}_k\}$, and an optional set of masks $M = \{\bm{m}_1, \cdots, \bm{m}_k\}$ corresponding to each $\bm{y}_i$, such an image editor $\mathcal{M}(\bm{I}_i, \bm{y}_i, \bm{m}_i)$ should be able to make the semantic modification intended by $\bm{y}_i$ on $\bm{I}_i$ iteratively. If the mask $\bm{m}_i$ is provided by the user, then the model $\mathcal{M}$ should constrain the edits to follow $\bm{m}_i$. The set of edited images $\mathcal{I}_{edits} = \{\bm{I}_1, \cdots, \bm{I}_k\}$ should be visually appealing and semantically consistent to $E$ and $M$.

Here, we propose one instantiation for $\mathcal{M}(\cdot, \cdot, \cdot)$ that builds on top of a pre-trained diffusion model \cite{brooks2022instructpix2pix} and does not need any retraining, instead relying only on test-time optimization of the intermediate representation of the model. The practicality of our newly introduced problem setting combined with the versatility of diffusion models allows \ours to be a faithful co-pilot for image editing workflows.
We explain how \ours handles multi-granular control and iterative edits in \cref{sec:multi_granular} and \cref{sec:iterative_editing} respectively. We explain our overall framework in \cref{sec:overall_framework}.


\begin{figure}[t]
  \centering
\includegraphics[width=1\columnwidth]{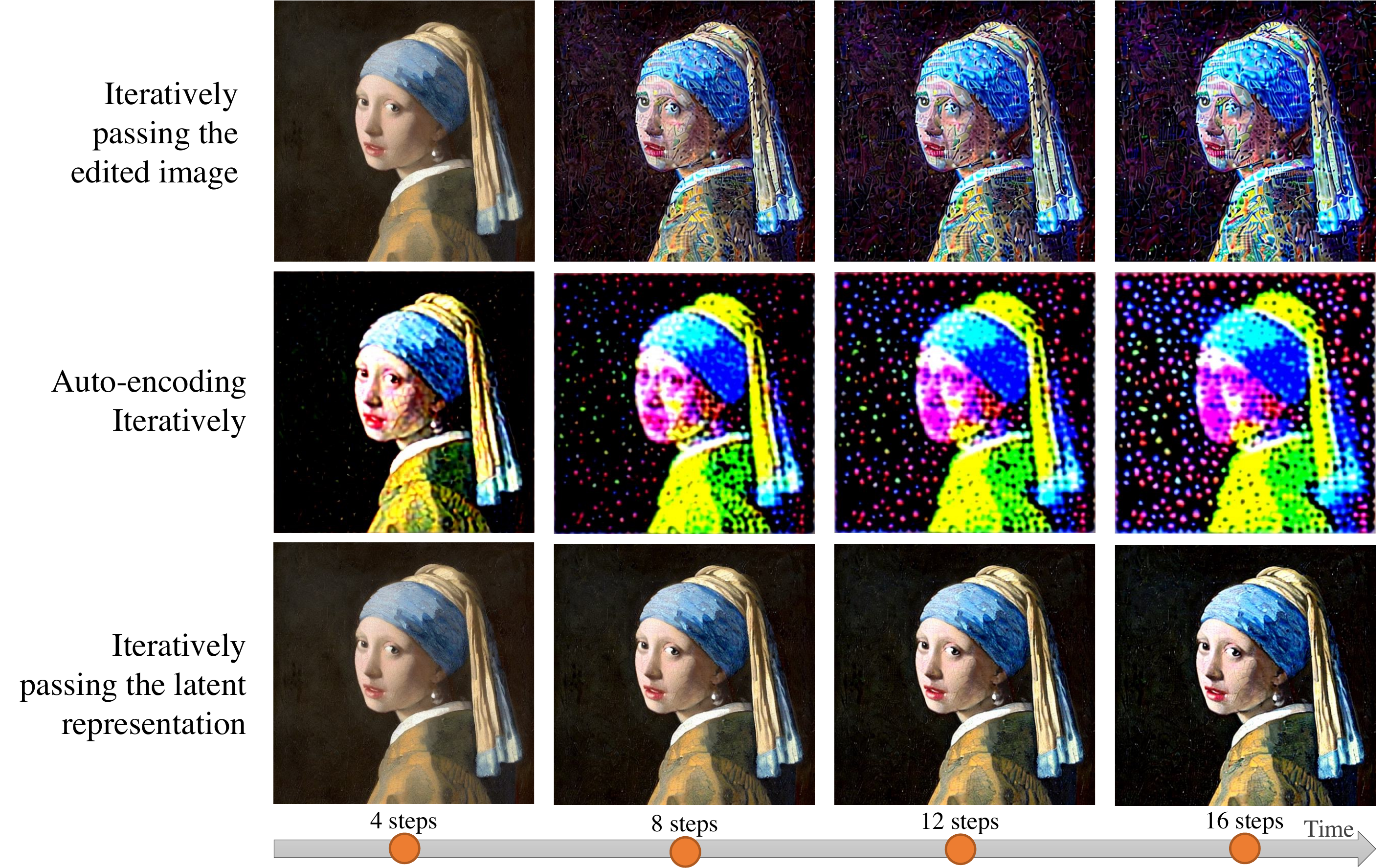}
  \caption{While iteratively passing an image through the diffusion model \cite{brooks2022instructpix2pix}, we see noisy artifacts being added in successive steps (we illustrate $4, 8, 12 \text{ and } 16$ steps) in Row 1. We see a similar phenomena while only auto-encoding the same image iteratively (here, the diffusion model is not used) in Row 2. Finally, when we iterate in the latent space (where $\bm{z}_{img}^{e+1}$ is passed iteratively), we see that the successive images are robust to such noisy artifacts. 
  }
  \label{fig:motivating_latent_iteration}
\end{figure}

\begin{figure*}
  \centering
\includegraphics[width=2\columnwidth]{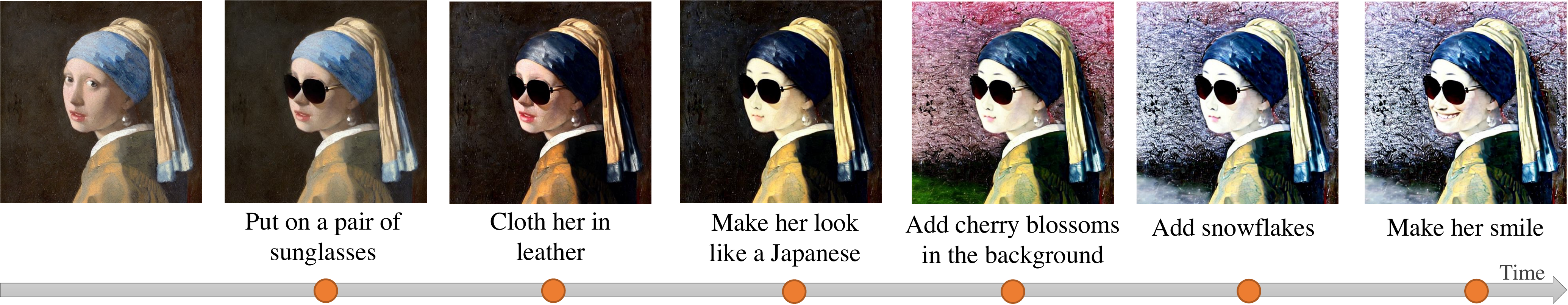}
  \caption{The figure shows how our proposed \textit{latent iteration} framework is able to semantically modify the first image, consistent with the text caption provided by the user at each time step. Please see \cref{sec:iterative_editing} for more details.
  }
  \label{fig:iterative_result}
  \vspace{-10pt}
\end{figure*}
\subsection{Multi-granular Image Editing}\label{sec:multi_granular}


The first contribution of our work is to equip diffusion models with the flexibility to constrain where an edit should be applied spatially on an image. We propose to interpret a diffusion model as an energy-based model (EBM) \cite{lecun2006tutorial,liu2022compositional}, leading to a flexible new method to selectively restrict gradient updates to the region of interest to the user. 

We first begin with a brief review of EBMs. An EBM provides a flexible way of modelling data likelihood. The probability density for the latent embedding $\bm{z} = \mathcal{E}(\bm{x})$, corresponding to an image $\bm{x}$ can be expressed as follows:
\begin{equation}
    p_{\psi}(\bm{z}) = \frac{\text{exp~}(-E_\psi(\bm{z}))}{\int_{\bm{z}} \text{exp~}(-E_\psi(\bm{z})) d\bm{z}},  \label{eqn:ebm}
\end{equation}
where $E_\psi(\cdot)$ is an energy function which maps $\bm{z}$ to a single scalar value, called the energy or score. $E_\psi(\cdot)$ is usually instantiated as a neural network with parameters $\psi$. After learning, sampling from the EBM is indeed challenging, owing to the intractability of the partition function ($\int_{\bm{z}} \text{exp~}(-E_\psi(\bm{z})) d\bm{z}$). A popular MCMC algorithm called Langevin Sampling \cite{neal2011mcmc, welling2011bayesian}, which makes use of the gradient of  $E_\psi(\cdot)$ is used as the surrogate as follows:

\begin{equation}
    \mathbf{z}_{t} = \mathbf{z}_{t-1} - \frac{\lambda}{2}\partial_{\mathbf{z}} E_{\psi}(\mathbf{z_{t-1}}) + \mathcal{N}(0, \omega_t^2\mathbf{I})
\label{eqn:ebm_sampling}
\end{equation}
where $\lambda$ is step size and $\omega$ captures the variance of samples.

Interestingly, the sampling process used by the EBM in \cref{eqn:ebm_sampling} and that used by the latent diffusion model in \cref{eqn:inference} are functionally same. Without loss of generality, this allows us to express the \textit{noise predictions} from the diffusion model $\epsilon_\theta(\cdot, \cdot)$ as the \textit{learned gradients} of the energy function $E_\psi(\cdot)$ \cite{liu2022compositional}. Thus we can control the granularity of each edits by selectively zeroing out the gradients (equivalent to masking parts of the noise predictions) that we are not interested in updating. This theoretically grounded approach turns out to be simple to implement. 
For a user provided mask $\bm{m} \in \{0, 1\}^{w\times h \times 1}$, which specifies the location to constrain the edits, we update the iterative denoising step in \cref{eqn:inference} as follows:
\begin{equation}
    \bm{z}_{t-1} = \bm{z}_{t} - \bm{m} * \epsilon_\theta(\bm{z}_t, t) + \mathcal{N}(\bm{0}, \sigma_t^2\textbf{I}) \label{eqn:inference_with_mask}
\end{equation}
Our experimental analysis in \cref{sec:results}, shows that such gradient modulation is effective in localizing the user intent (conveyed via the binary mask), without adding any extra training or compute expense. 


\subsection{Iterative Image Editing} \label{sec:iterative_editing}

The next contribution of our work is to iteratively edit images with diffusion models while maintaining the state of the canvas, i.e., the newer edit instruction from a user should be applied on the latest version of the image.

\begin{figure}[H]
  \centering
\includegraphics[width=1\columnwidth]{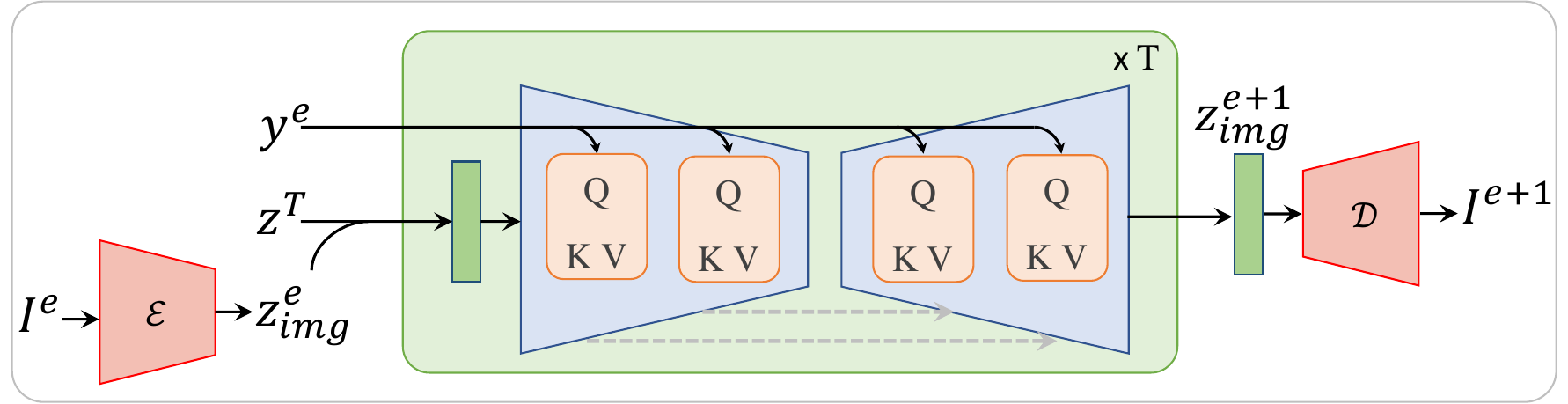}
  \caption{The figure illustrates the architectural components \cite{brooks2022instructpix2pix} involved in editing an image $\bm{I}^e$, according to an edit instruction $\bm{y}^e$ in the $e^{th}$ edit iteration. $\bm{z}^{T} \sim \mathcal{N}(\bm{0},\textbf{I})$ is successively denoised by the diffusion model for $T$ iterations to generate $\bm{z}_{img}^{e+1}$. 
  }
  \label{fig:forward_pass}
\end{figure}
Diffusion models capable of doing image editing \cite{brooks2022instructpix2pix,wang2022imagen} can consume the image to be edited. We briefly describe its architectural components in \cref{fig:forward_pass}.
Consider $\bm{y}^e$ to be the edit instruction that needs to be applied to an image $\bm{I}^e$, at an edit step $e$. 
$\bm{I}^e$ is passed though a pretrained VQ-VAE encoder $\mathcal{E}$ to obtain $\bm{z}_{img}^e$.
$\bm{z}_t \sim \mathcal{N}(\bm{0},\textbf{I})$ is the initial latent variable.
$\bm{z}_{img}^e$ is stacked with $\bm{z}_t$ and  successively denoised by the pretrained diffusion model (shown in the light green outline) over $T$ iterations.

A naïve approach to iteratively edit the image would be to pass the edited image $\bm{I}^{e+1}$ through the model along with the next edit instruction $\bm{y}^{e+1}$. Unfortunately, this accumulates and amplifies the noisy artifacts in the image. The first row of \cref{fig:motivating_latent_iteration} illustrates this phenomena. In this experiment, we iteratively pass an image through the architecture defined in \cref{fig:forward_pass}, and use $\bm{y}^e = \phi$ (this is to characterize the behaviour independent of the edit instruction at each step).

On a high level, there would potentially be only two sources that might introduce such noisy artifacts: 1) the VQ-VAE based auto-encoder and 2) the denoising steps of the latent diffusion model. In-order to isolate each of their contribution, we experiment using only the VQ-VAE based encoder and decoder from the pipeline described in \cref{fig:forward_pass}; \ie $\bm{I}^{e+1} = \mathcal{D}(\mathcal{E}(\bm{I}^{e}))$. We show the corresponding results in the second row of \cref{fig:motivating_latent_iteration}. These result illustrates that the auto-encoding is not perfect, and is indeed accumulating noise over time. 
This motivates us to propose \textit{latent iteration}, where the latent representation produced by diffusion model corresponding to edit instruction $\bm{y}^{e}$ will be passed iteratively along with $\bm{y}^{e+1}$ in successive edit step. Hence, $\bm{z}_{img}^{e+1}$ will be used instead of $\mathcal{E}(\bm{I}^{e+1})$ as input to diffusion model. This simple approach alleviates the exacerbation of the noisy artefacts as seen in the last row of \cref{fig:motivating_latent_iteration}.

\begin{figure}[h]
  \centering
\includegraphics[width=1\columnwidth]{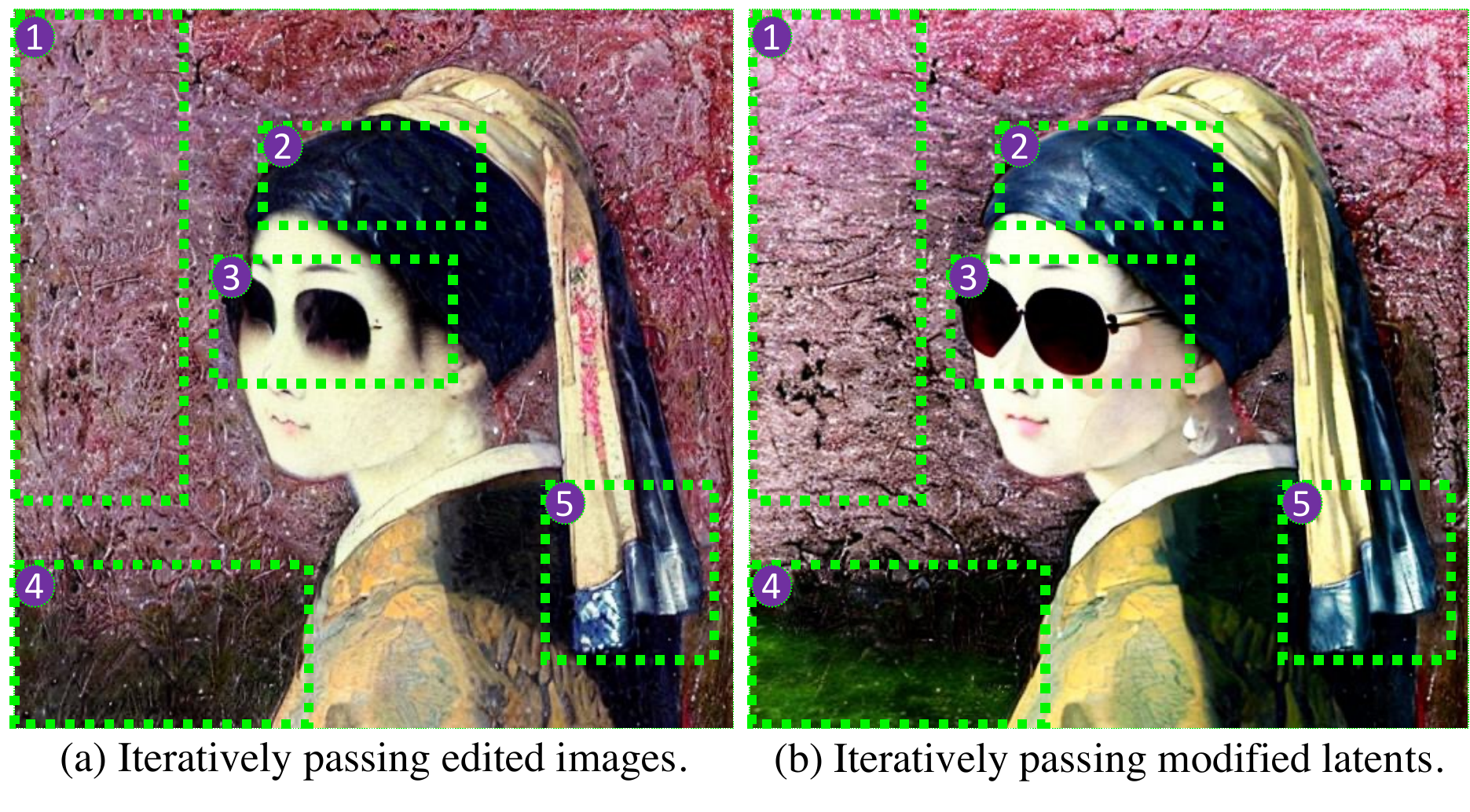}
  \caption{The figure compares the fourth edit step in \cref{fig:iterative_result} between image space and latent space iteration. Green boxes $2, 3 \text{ and } 5$ shows improved preservation of semantic concepts from the previous edit step, while box $4$ shows how newer related concepts are added, while keeping the image less noisy (box $1$). Kindly zoom in for fine details.
  }
  \label{fig:comparison}
\end{figure}

In \cref{fig:iterative_result}, we apply latent iteration over multiple edit instructions $\bm{y}^e$. We note that our method is able to make semantic edits consistent with the corresponding caption in each edit iteration. 
Further, in \cref{fig:comparison}, we compare the fourth edit result between iterating over image space and latent space. (We showcase the results of the other steps in the Supplementary owing to space constrains.) 
It is interesting to note that latent iteration is not only able to reduce the noise accumulation (box $1$), but also improve the consistency with previous edits. In box $3$, the \texttt{sunglass} from the previous step is retained as is (note the frame), while in box $2$ and $5$, the \texttt{leather} head cover is maintained with the newer edits. In box $4$, latent iteration was able to add related semantic concept to improve the consistency of the image. 
We see that this behaviour consistently holds across our exhaustive experimental evaluation in \cref{sec:experiments}.

\subsection{Overall Framework} \label{sec:overall_framework}
\begin{algorithm}
\small
\caption{It\underline{e}rative \underline{M}ult\underline{i}-granu\underline{l}ar \underline{I}mage \underline{E}ditor}
\label{algo:emilie}
\begin{algorithmic}[1]
\Require{Image to be Edited: $\bm{I}_0$; Edit Instructions: $E = \{\bm{y}_1,\cdots,\bm{y}_k\}$; 
Optional Masks: $M = \{\bm{m}_1,\cdots,\bm{m}_k\}$; Pretrained Diffusion Model \cite{brooks2022instructpix2pix}: $\epsilon_\theta(\cdot,\cdot,\cdot)$; VQ-VAE: $\mathcal{E}(\cdot), \mathcal{D}(\cdot)$; Variance Schedule: $\sigma_t$; Number of Diffusion Steps: $T$.
}
\Ensure{Edited Images: $\mathcal{I}_{edits}$}
\State $\bm{z}_{init} \sim \mathcal{N}(\bm{0}, \textbf{I})$ 
\For{$\bm{y}_e \in E$} \Comment{\textit{For each edit instruction.}}
\If{e == $1$}
\State $\bm{z}_{img}^e \leftarrow \mathcal{E}(\bm{I}_0)$
\Else
\State $\bm{z}_{img}^e \leftarrow \text{prev\_latent}$ \Comment{\textit{Latent Iteration.}}
\EndIf
\State $\bm{z}_{T} \leftarrow \text{concat}(\bm{z}_{init}, \bm{z}_{img}^e)$ 
\For{$t \in \{T, \cdots, 0\}$} \Comment{\textit{For each denoising step.}}
\State $\bm{z}_{t-1} = \bm{z}_{t} - \bm{m}_e * \epsilon_\theta(\bm{z}_t, t, \bm{y}_e) + \mathcal{N}(\bm{0}, \sigma_t^2\textbf{I}) $ \Comment{\textit{\cref{eqn:inference_with_mask}}}
\EndFor
\State $\mathcal{I}_{edits}.\text{append}(\mathcal{D}(\bm{z}_0))$
\State prev\_latent = $\bm{z}_0$
\EndFor
\State \Return $\mathcal{I}_{edits}$
\end{algorithmic}
\end{algorithm}

\cref{algo:emilie} summarizes the key steps involved in adding multi-granular edits to an image iteratively. For the first edit instruction $\bm{y}_1$, we initialize $\bm{z}_{img}^e$ to the VQ-VAE encoding of the input image $\bm{I}_0$ in Line $4$. This latent is therein passed to the diffusion model via $\bm{z}_T$ in Line $7$. If the user specifies a mask to control the spatial reach of the edit, it is used while denoising the latent in Step $9$. After denoising $\bm{z}_0$ is indeed decoded using VQ-VAE decoder, and stored to $\mathcal{I}_{edits}$. Importantly, $\bm{z}_0$ is saved and reused for subsequent edit instructions in Line $11$ and Line $6$ respectively.

\section{Experiments and Results} \label{sec:experiments}
\subsection{\ourbenchmark Benchmark}
To complement our newly introduced problem setting, we introduce \ourbenchmark (\underline{I}terative \underline{M}ulti-granular \underline{I}mage \underline{E}diting \underline{Bench}mark) to evaluate the efficacy of the methodologies. Inspired by TEdBench \cite{kawar2022imagic}, we manually curate $40$ images from LAION \cite{schuhmann2022laion} dataset. These include images of people, landscapes, paintings, and monuments. Next, we collected four semantically consistent edit instructions that would modify these images. This would help to quantify the quality of the iterative edits by the model. We also add masks to some of these edit instructions to simulate local editing. We hope that \ourbenchmark would serve as a standardized evaluation setting for this task.

\begin{figure*}
  \centering
  \vspace{-10pt}
\includegraphics[width=1.9\columnwidth]{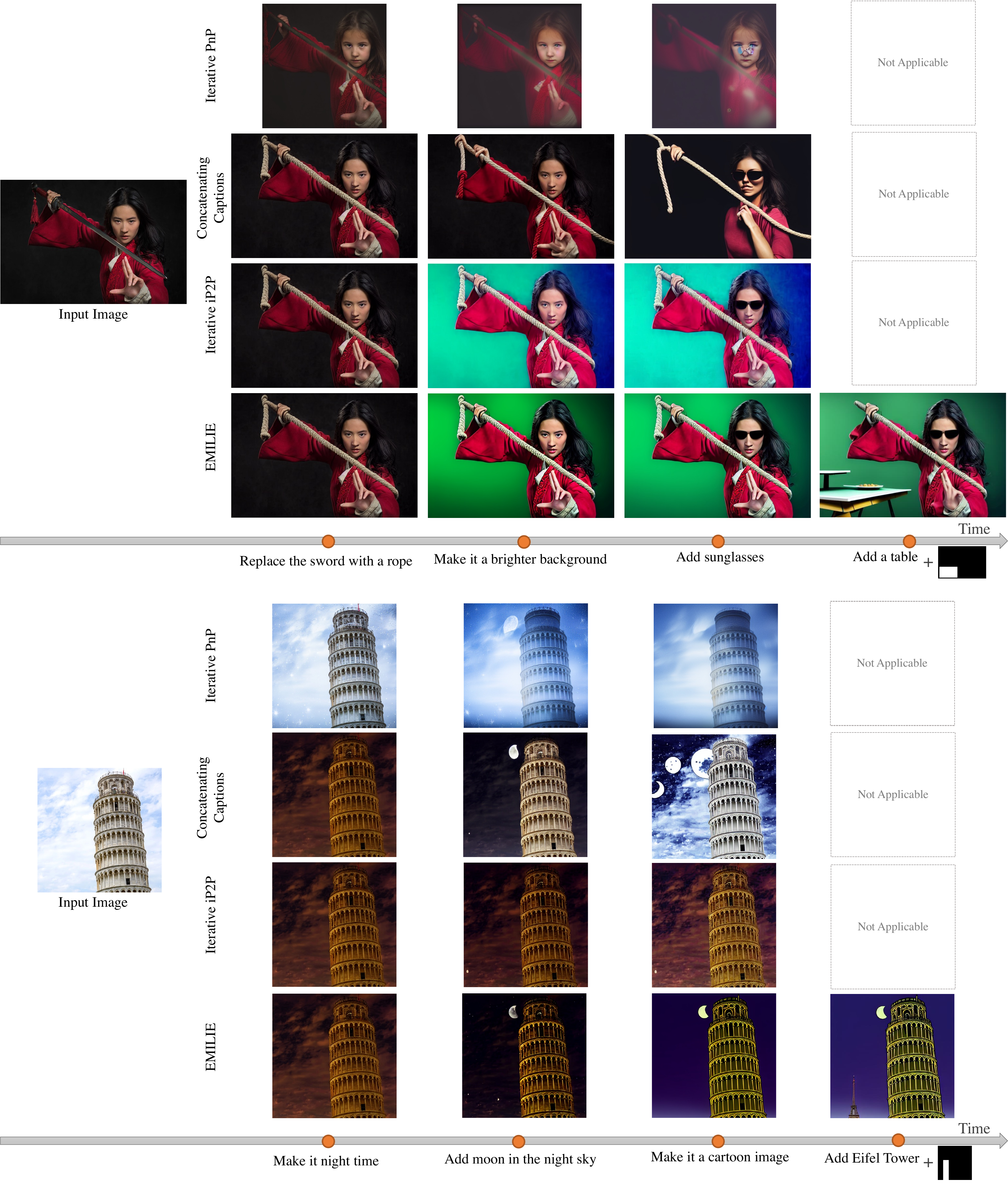}
  \caption{Here we compare \ours with three baselines on images from \ourbenchmark. `Iterative PnP' and `Iterative iP2P' refers to Plug and Play \cite{tumanyan2022plug} and Instruct Pix2Pix \cite{brooks2022instructpix2pix}, where the latest edited image is recursively passed on in the next edit step. `Concatenating Captions' refers to the setting where all the edit instructions that have been received so far, is concatenated and send through Instruct Pix2Pix \cite{brooks2022instructpix2pix} to edit the original image. We see that \ours is able to consistently reduce the noisy artefact accumulation and retain semantics of earlier steps better. Please see more results in the Supplementary material.
  }
  \label{fig:qualitative_results}
  \vspace{-10pt}
\end{figure*}
\subsection{Experimental Protocol}\label{sec:protocol}
We evaluate the `iterative' and `multi-granular' aspects of our approach both qualitatively and quantitatively. We use examples from \ourbenchmark and compare against three baseline approaches. Two of these include iteratively passing the edited image through two image editing methods: InstructPix2Pix \cite{brooks2022instructpix2pix} and Plug and Play \cite{tumanyan2022plug}. These methods are representative of two kinds of approaches for diffusion-based image editors. Plug and Play uses DDIM inversion to project the image to the latent space of the diffusion model, while InstructPix2Pix learns a few new layers to directly pass the input image into the model. The third baseline is to concatenate all the instructions received until a time step, and pass it through InstructPix2Pix along with the original image. We show these results in the following sections.

Further, to better understand the multi-granular local editing aspect of \ours, we compare with two recent state-of-the-art mask-based image editing methods:
DiffEdit \cite{couairon2023diffedit} and Blended Latent Diffusion \cite{avrahami2022blended_latent} on the EditBench \cite{wang2022imagen} dataset. We could not compare with Imagen Editor \cite{wang2022imagen} because their code, models or results are not publicly available. We use two complementary metrics for quantitative evaluation: first, we compute the CLIP \cite{radford2021learning} similarity score of the edited image with the input edit instruction and the second is a text-text similarity score, where BLIP \cite{li2022blip} is used to first caption the generated image, and it is then compared with the edit instruction.

\subsection{Implementation Details}
We build \ours by extending the InstructPix2Pix diffusion model \cite{brooks2022instructpix2pix} to support iterative multi-granular editing capability. We keep all the hyper-parameters involved to be same as their implementation. We use a single NVIDIA A-100 GPU to do our inference. Each edit instruction takes $6.67$ seconds on average to complete. 
While doing latent iteration, we find that the magnitude of values in 
$\bm{z}_{img}^{e+1}$  is significantly lower than $\mathcal{E}(\bm{I}^{e+1})$. To normalize this, we multiply $\bm{z}_{img}^{e+1}$ with a factor $f = avg(\mathcal{E}(\bm{I}^{e+1})) / avg(\bm{z}_{img}^{e+1})$. We use ancestral sampling with Euler method to sample from the diffusion model.

\subsection{Results on \ourbenchmark} \label{sec:results}
We showcase our qualitative results in \cref{fig:qualitative_results}. 
We compare with iterative versions of InstructPix2Pix \cite{brooks2022instructpix2pix} and Plug and Play (PnP) \cite{tumanyan2022plug} and a concatenated set of edit instructions, on images from \ourbenchmark. 

While analyzing the results, we note that \ours is able to retain the concepts from the previous edits, without deteriorating the quality in successive iterations. Iterative PnP struggles the most. This can be attributed to the DDIM inversion which projects the image to the latent space of the diffusion model. The concatenated caption (that contains all the edit instruction that we have seen so far) contains multiple concepts. The diffusion model tries its best to generate these multi-concept images, but struggles to maintain consistency with the previously edited version of the  image. Iterative InstructPix2Pix performs the closest to \ours, but accumulates noise as edits progresses.

\ours supports multi-granular edits too, as is shown in the last column. The baseline methods cannot operate in this setting. We compare \ours with multi-granular approaches in \cref{sec:editbenchResults}. These results illustrate that iterating in latent space indeed helps attenuate noise, and gives the model the optimal plasticity to add new concepts yet retain the stability to not forget the previous set of edited concepts.

\begin{figure}[h]
  \centering
\includegraphics[width=1\columnwidth]{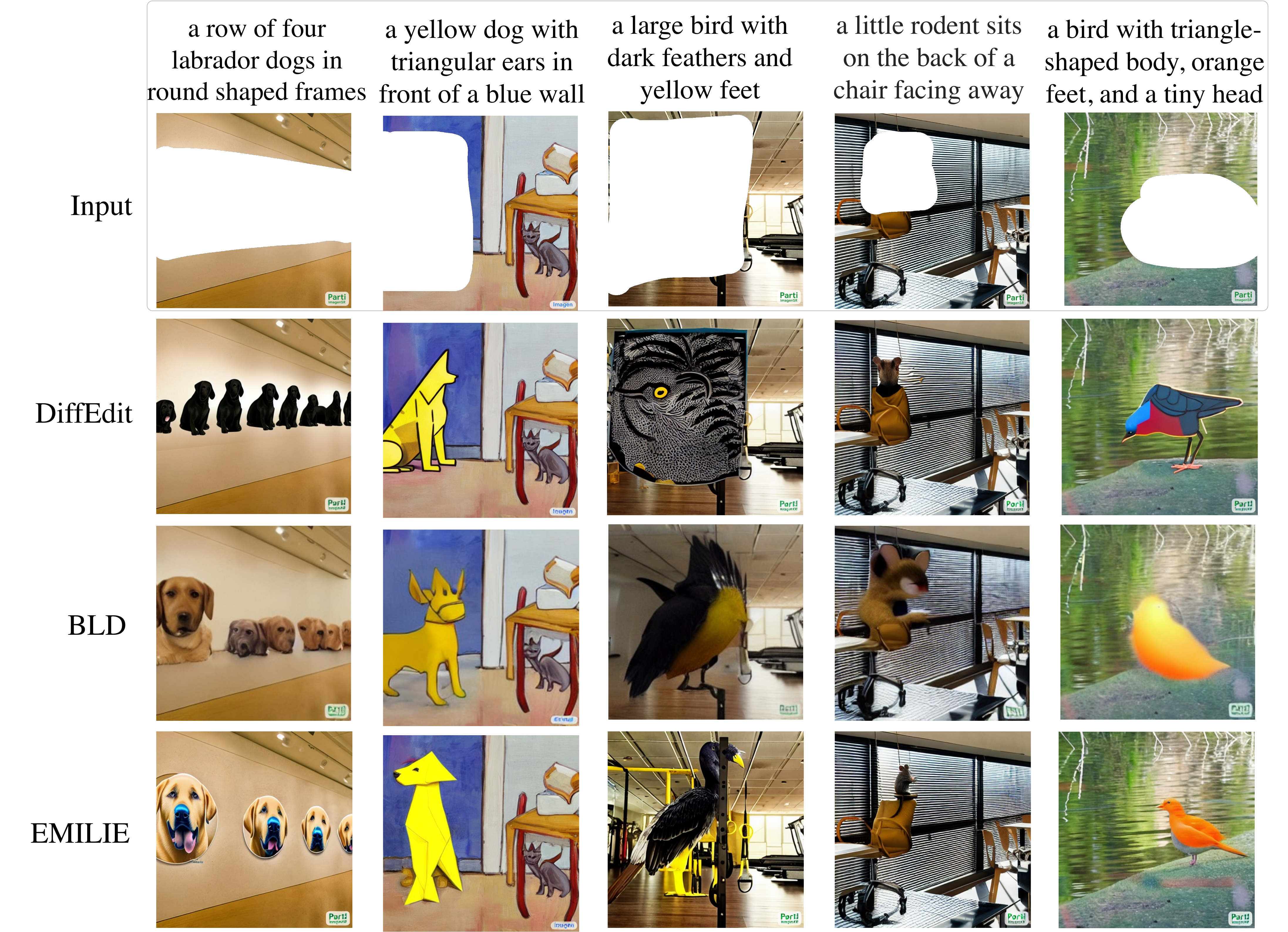}
  \caption{We illustrate the ability of \ours to do local editing on images from EditBench \cite{wang2022imagen} benchmark. We compare against recent state-of-the-art methods, DiffEdit (ICLR `23) \cite{couairon2023diffedit} and Blended Latent Diffusion (SIGGRAPH `23) \cite{avrahami2022blended_latent}. Our simple gradient modification approach is able to consistently improve the quality of generation when compared to these approaches.
  }
  \label{fig:local_edits}
\end{figure}

\subsection{Results on EditBench}\label{sec:editbenchResults}
\cref{fig:local_edits} shows the results of using \ours for doing localized edits in images. The top row includes the input image with mask and the corresponding edit text from the user. The subsequent two rows include results from two recent state-of-the-art approaches: DiffEdit \cite{couairon2023diffedit} and Blended Latent Diffusion \cite{avrahami2022blended_latent}, followed by \ours.
The results show that the methods are successful in limiting the extent of edits to the area constrained by the user. \ours is able to make semantically richer modifications to the masked region. Our training-free guidance during the denoising steps is able to constrain the edits to local regions, while being more semantically consistent. 

Finally, we run a quantitative evaluation of how well the edited image is able to capture the semantics of the edit instruction successfully by measuring the CLIP and BLIP scores. The results are provided in \cref{tab:clipblip}. We comfortably outperform the baselines here too.

\begin{table}[h]
\caption{We quantitatively evaluate the performance of \ours for multi-granular editing here. When compared to recent state-of-the-art approaches, \ours is able to score better performance in both CLIP and BLIP score metrics.}
\label{tab:clipblip}
\resizebox{0.47\textwidth}{!}{%
\begin{tabular}{@{}r|ccccc@{}}
\toprule
                   & \multicolumn{2}{c}{DiffEdit \cite{couairon2023diffedit}} & \multicolumn{2}{c}{BLD \cite{avrahami2022blended_latent}} & \ours \\ \midrule
Average CLIP Score & 0.272         & {\color{red}(-0.039)}        & 0.280       & {\color{red}(-0.031)}      & \textbf{0.311}  \\
Average BLIP Score & 0.582         & {\color{red}(-0.038)}        & 0.596      & {\color{red}(-0.024)}      & \textbf{0.620}   \\ \bottomrule
\end{tabular}%
}
\end{table}

\subsection{User Study}
We conduct a user study with images from \ourbenchmark and EditBench \cite{wang2022imagen} datasets to test the preference of users across the generated images in \cref{tab:user_iterative} and \cref{tab:user_mg} respectively. The former evaluates iterative edits, while the latter analyzes the preference for multi-granular edits. From the $30$ users, most of them preferred the generations from \ours over the other baselines.

\begin{table}
\vspace{5pt}
\parbox{.45\linewidth}{
\centering
\caption{User Study for Multi-granular Edits.}
\begin{tabular}{@{}l|c@{}}
\toprule
   Method      & Split \\ \midrule
DiffEdit \cite{couairon2023diffedit} &   4.87\%    \\ 
BLD \cite{avrahami2022blended_latent}      &    13.33\%   \\
\ours   &   81.79\%    \\ \bottomrule
\end{tabular}%
\label{tab:user_mg}
}
\hfill
\parbox{.45\linewidth}{
\centering
\caption{User Study for Iterative Edits.}
\begin{tabular}{@{}l|c@{}}
\toprule
   Method      & Split \\ \midrule
Concat &    18.18 \%  \\
iP2P      &   19.32 \%    \\
\ours   &   62.50 \%   \\ \bottomrule
\end{tabular}%
\label{tab:user_iterative}
}
\end{table}

\section{Further Discussions and Analysis}
\subsection{Object Insertion with \ours}
Our proposed approach gives control to the user in specifying the location of edits. At times, users might have an image of an object that they would like to insert into a base image. It would be great if the model can automatically propose a plausible location and then insert the object there. We find a straight forward strategy to combine \ours with GracoNet \cite{zhou2022learning} (a recent method for proposing object placement in images) to achieve this. Given a base image and an inset image, GracoNet proposes masks for potential location for the inset image. Next, we pass the inset image thought BLIP 2 \cite{li2022blip} model to generate a caption. Finally, the mask, the caption and base image is passed to \ours to render the final image.

\begin{figure}
  \centering
\includegraphics[width=1\columnwidth]{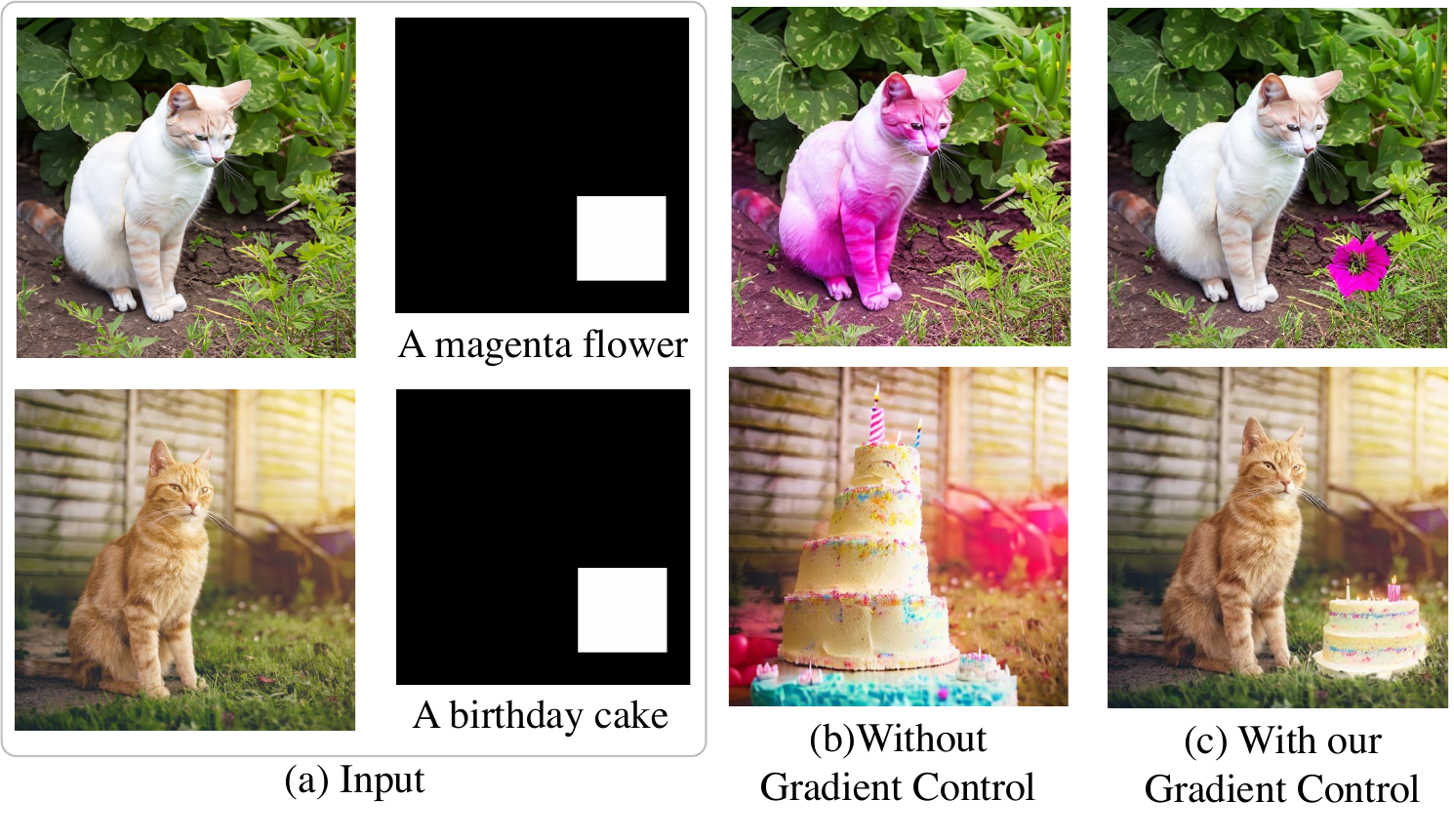}
  \caption{Here we experiment with turning off the gradient control strategy explained in \cref{sec:multi_granular} for local edits. Without modulating the gradients, the characteristics of the edit instruction gets globally applied on the image, whereas \ours is able to easily localise the edits effectively.
  }
  \label{fig:graconet}
\end{figure}

\cref{fig:graconet} showcases some examples from OPA dataset \cite{liu2021opa}. We can see that the object insertion by \ours is more composed and realistic than that by GracoNet. This is because, instead of just placing the object at a location, \ours indeed synthesises a new object at the location specified via the mask proposed by GracoNet. 

\begin{figure}
  \centering
\includegraphics[width=1\columnwidth]{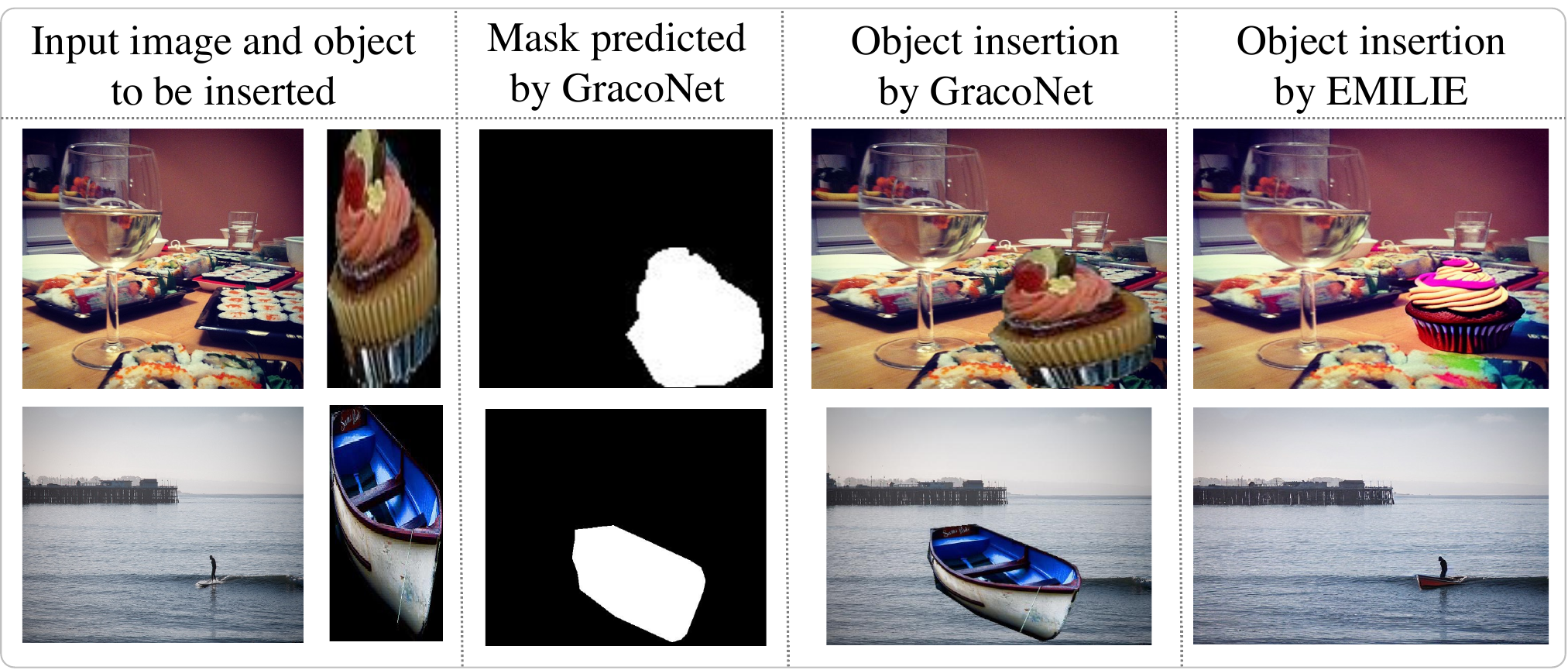}
  \caption{We re-purpose the local editing capability of \ours to insert objects into specific locations of a source image. For this, we combine \ours with GracoNet \cite{zhou2022learning} which predicts the location of the object to be inserted. We can see from the results that the insertions done by \ours is more composed to the background. 
  }
  \label{fig:ablation}
\end{figure}

\subsection{Ablating the Gradient Control}
In order to understand the contribution of our proposed gradient control strategy explained in \cref{sec:multi_granular}, we turn it off while doing local editing. Results in \cref{fig:ablation} shows that without modulating the latent representations selectively (following \cref{eqn:inference_with_mask}), the edit instructions gets applied globally.

\subsection{Limitations}
While experimenting with \ours, we could understand that it cannot handle negative edit instructions. Let us say we add a pair of sun-glasses as the first edit, and try removing it in the second edit step, the model fails to give back the original image. Disentangling the feature representations for each edits would potentially help to alleviate this issue. We will explore this in future work. We show more failure cases in the Supplementary materials.

\section{Conclusion}
We introduce a novel problem setting of \textit{Iterative Multi-granular Image Editing} where a creative professional can provide a series of edit instructions to be made to a real image. Optionally, they can specify the spatial locality on which the edit needs to be applied. Our proposed approach \ours is training free and utilizes a pre-trained diffusion model with two key contributions:
latent iteration (\cref{sec:iterative_editing}) for supporting iterative editing and gradient modulation (\cref{sec:multi_granular}) for supporting multi-granular editing. Finally, we introduce a new benchmark dataset \ourbenchmark, and bring out the mettle of our approach by comparing \ours against other state-of-the-art approaches adapted to our novel task. We hope that this newly identified research area would be actively investigated by the community.

{\small
\bibliographystyle{ieee_fullname}
\bibliography{egbib}
}

\clearpage
\appendix

\section{Alternate Iterative Editing Approaches}
\vspace{-5pt}
We would like to briefly discuss some alternate approaches that we had explored towards addressing iterative editing. Though intuitive, these approaches was found to be less effective than our proposed \textit{latent iteration} approach.

\subsection{Change Isolation and Feature Injection}

\begin{figure}[h]
  \centering
  \vspace{-15pt}
\includegraphics[width=1\columnwidth]{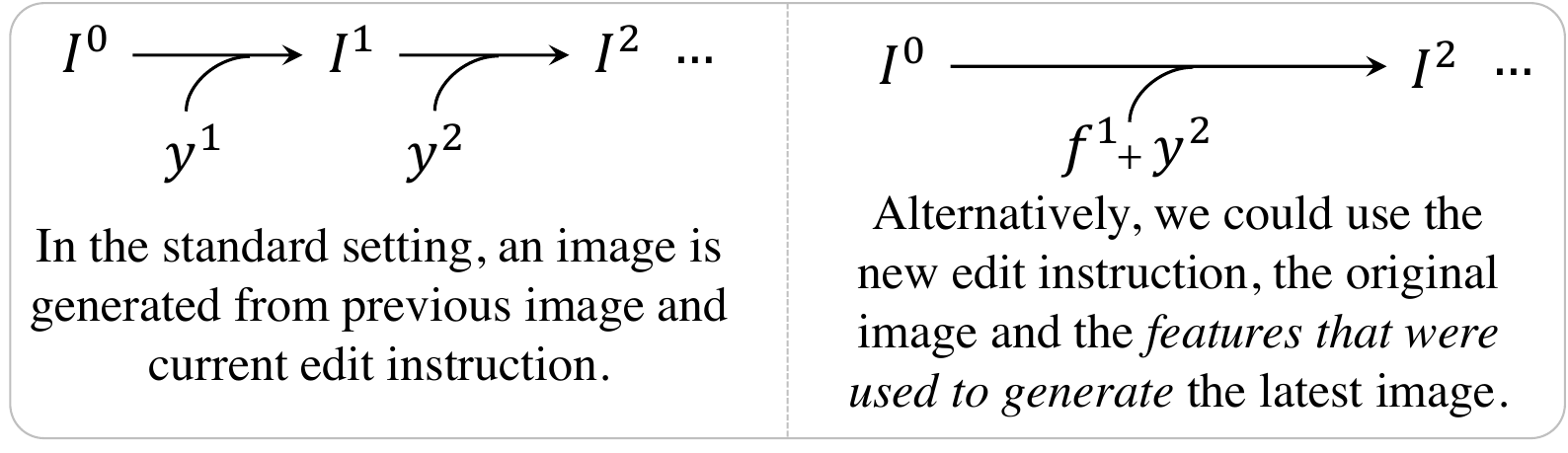}
  \vspace{-20pt}
  \caption{Ideally, it would be effective if we could isolate the changes intended by each edit instruction to the original image $\bm{I}_0$, and compose the final image by adding them successively to $\bm{I}_0$. We propose to do this in the feature space.
  }
    \vspace{-10pt}
\label{fig:feature_injection}
\end{figure}

Iterative editing involves making semantic changes corresponding to a set of edit instructions $\{\bm{y}_0, \cdots, \bm{y}_n\}$ to the original image $\bm{I}_0$. If we could isolate the changes each $\bm{y}_i$ cause to $\bm{I}_{i-1}$, we can cumulate these changes and apply them directly to $\bm{I}_0$. This would completely side-step the noisy artifact addition issue that surfaces when an image is recursively passed though the model for editing.

We explore an approach that does the change isolation in the image space and applies it back to the image in its feature space. Let $\bm{I}_i$ be the image generated following the edit instruction $\bm{y}_i$. We identify the changes caused by $\bm{y}_i$ by taking a difference $\bm{I}_i^{\Delta} = \bm{I}_i - \bm{I}_{i-1}$. Next, we isolate the features $\bm{f}_i^{\Delta}$ corresponding to $\bm{I}_i^{\Delta}$ (by forwarding passing $\bm{I}_i^{\Delta}$ through the model), and inject them into the model while consuming $\bm{y}_{i+1}$ and $\bm{I}_0$. As we want to preserve the overall image statistics of the original image, we choose to inject the self-attention features into the decoder layers of the UNet \cite{ronneberger2015u}, following Tumanyan \etal \cite{tumanyan2022plug} and  Ceylan \etal \cite{ceylan2023pix2video}. Though this approach helps to preserve the background of $\bm{I}_0$, the newer edits were not well represented in the images. The high-frequency elements like edges were well carried over, but finer details are mostly ignored.

\vspace{-5pt}
\subsection{Iterative Noise Removal}
Another approach is to explicitly reduce noise that gets accumulated while we iteratively pass the edited image through the model for successive edits. We used Gaussian blur towards this effort. Though this approach indeed reduces the noise accumulation, it significantly blurs the subsequent edit images. Feature injection to the self-attention layers were able to partially reduce the effect of blurring, but the images were not better than doing \textit{latent iteration}. 

\section{Latent Iteration vs. Image Iteration}

\begin{figure*}
  \centering
\includegraphics[width=1.75\columnwidth]{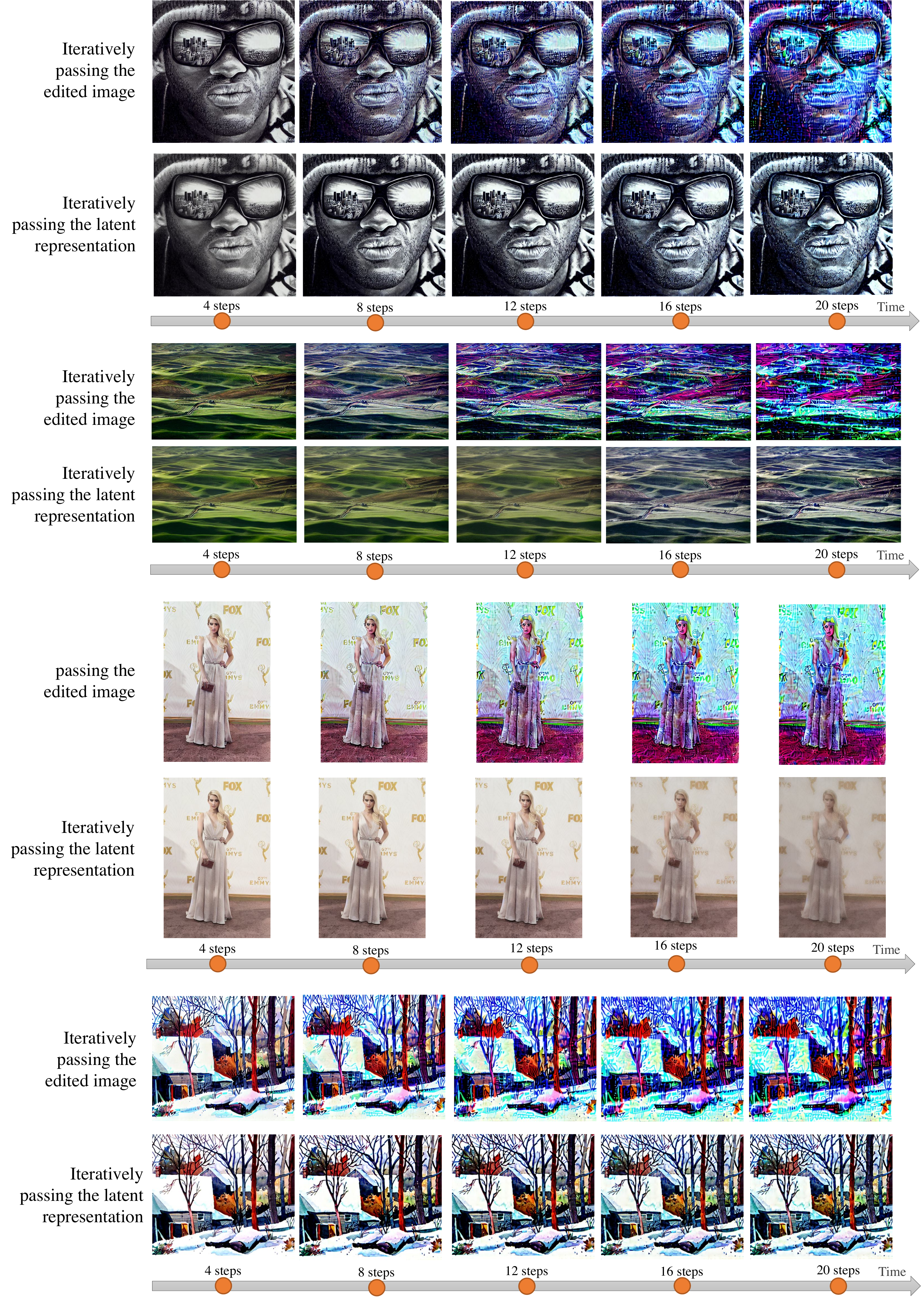}
  \caption{While iteratively passing an image through the diffusion model, we see noisy artifacts being accumulated (first row in each pair). Iterating over the latent representations helps minimise the artifact accumulation (second row in each pair).
  }
  \label{fig:suppl_results_noise}
  \vspace{-10pt}
\end{figure*}

\begin{figure*}
  \centering
\includegraphics[width=2\columnwidth]{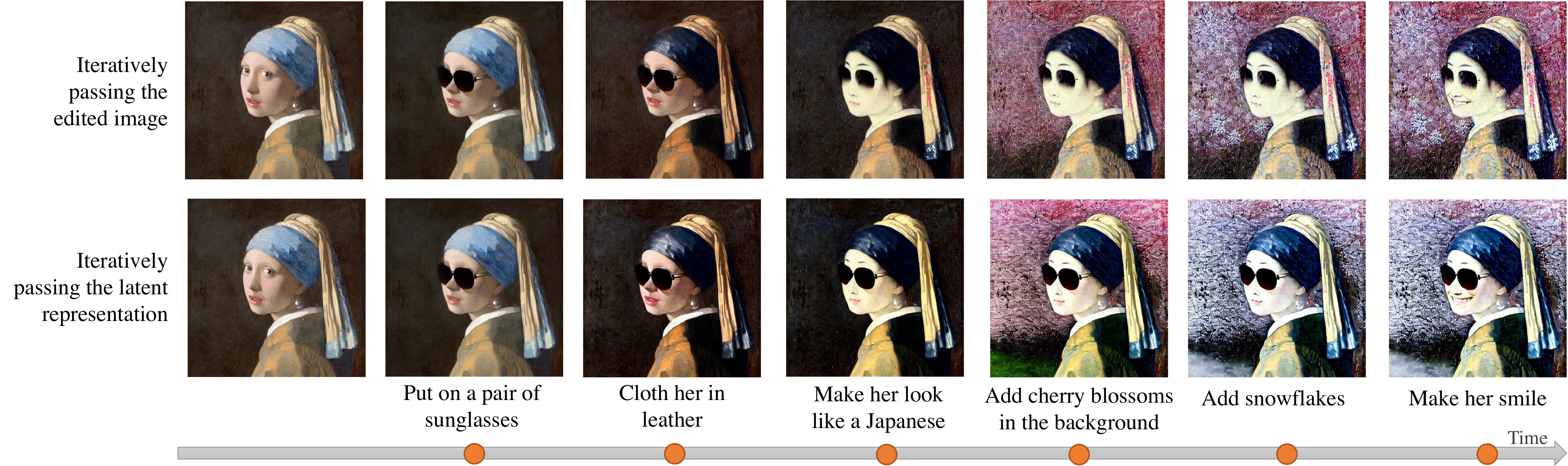}
  \caption{While iteratively editing images with a new edit instruction in each step, we observe that \ours is able to add new semantic information with minimal artifact accumulation.
  }
  \label{fig:earring}
  \vspace{-10pt}
\end{figure*}

A key finding of our analysis is quantifying the accumulation of noisy artifacts, while iteratively passing an image through the latent diffusion model. Here, in \cref{fig:suppl_results_noise}, we showcase a few more examples where we see degradation in the image quality while being iteratively processed. We choose photos, painting and landscape pictures for this study. We iteratively pass these images though the LDM (introduced in Sec. 4.2) for $20$ steps. We use a null string as the edit instruction to neutralize its contribution. As is evident from the figure, when we iterate in the image space, we see more severe image degradation when compared to iterating in the latent space. In \cref{fig:earring}, we compare with a new semantic edit instruction in each step.

\section{More Qualitative Comparisons}
\begin{figure*}
  \centering
\includegraphics[width=1.85\columnwidth]{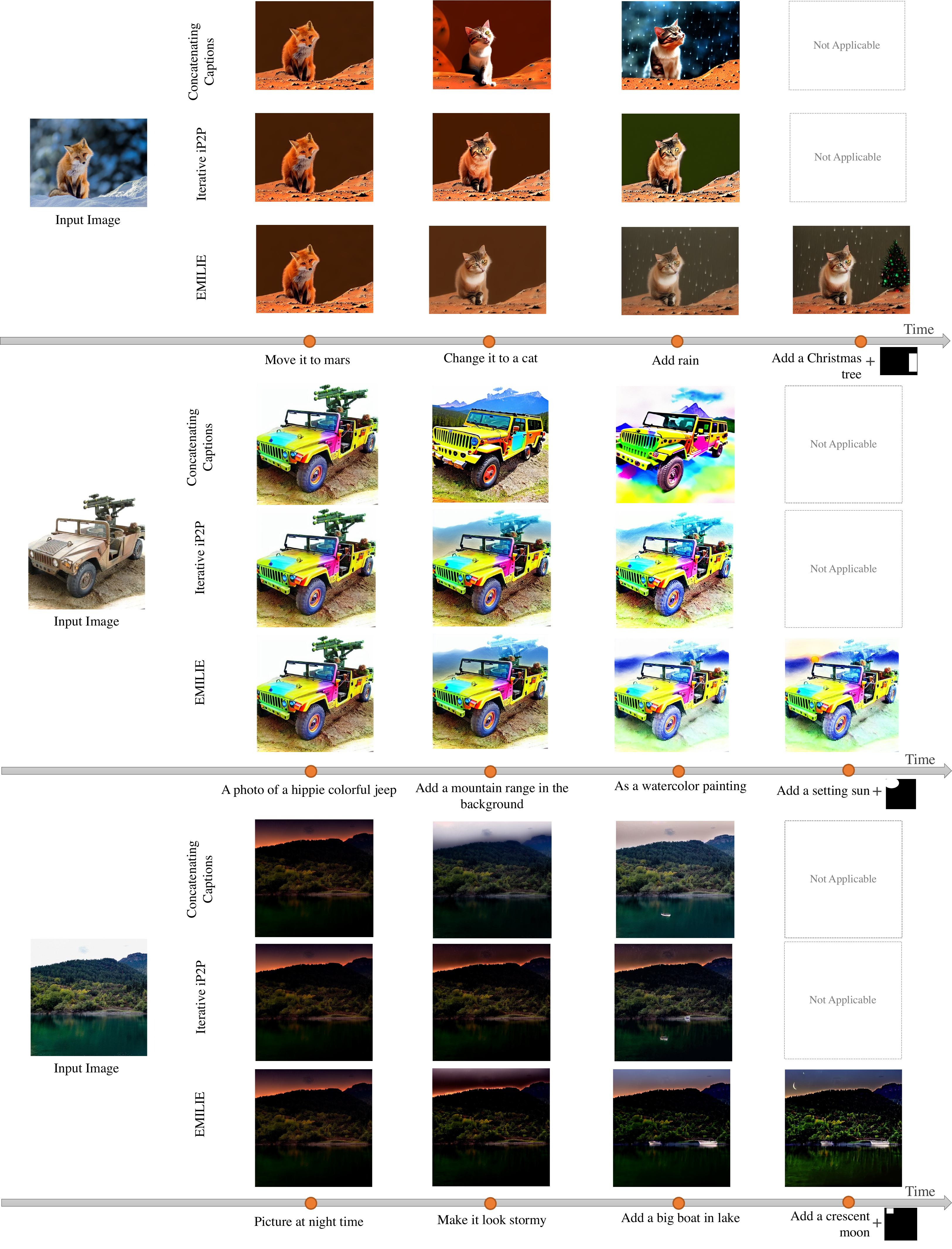}
  \caption{We showcase more qualitative comparisons with baselines. We see that \ours is able to produce more realistic and meaningful edits when compared to concatenating all the instructions seen so far, and iteratively using Instruct Pix2Pix \cite{brooks2022instructpix2pix}.
  }
  \label{fig:suppl_results_1}
\end{figure*}

\begin{figure*}
  \centering
\includegraphics[width=1.85\columnwidth]{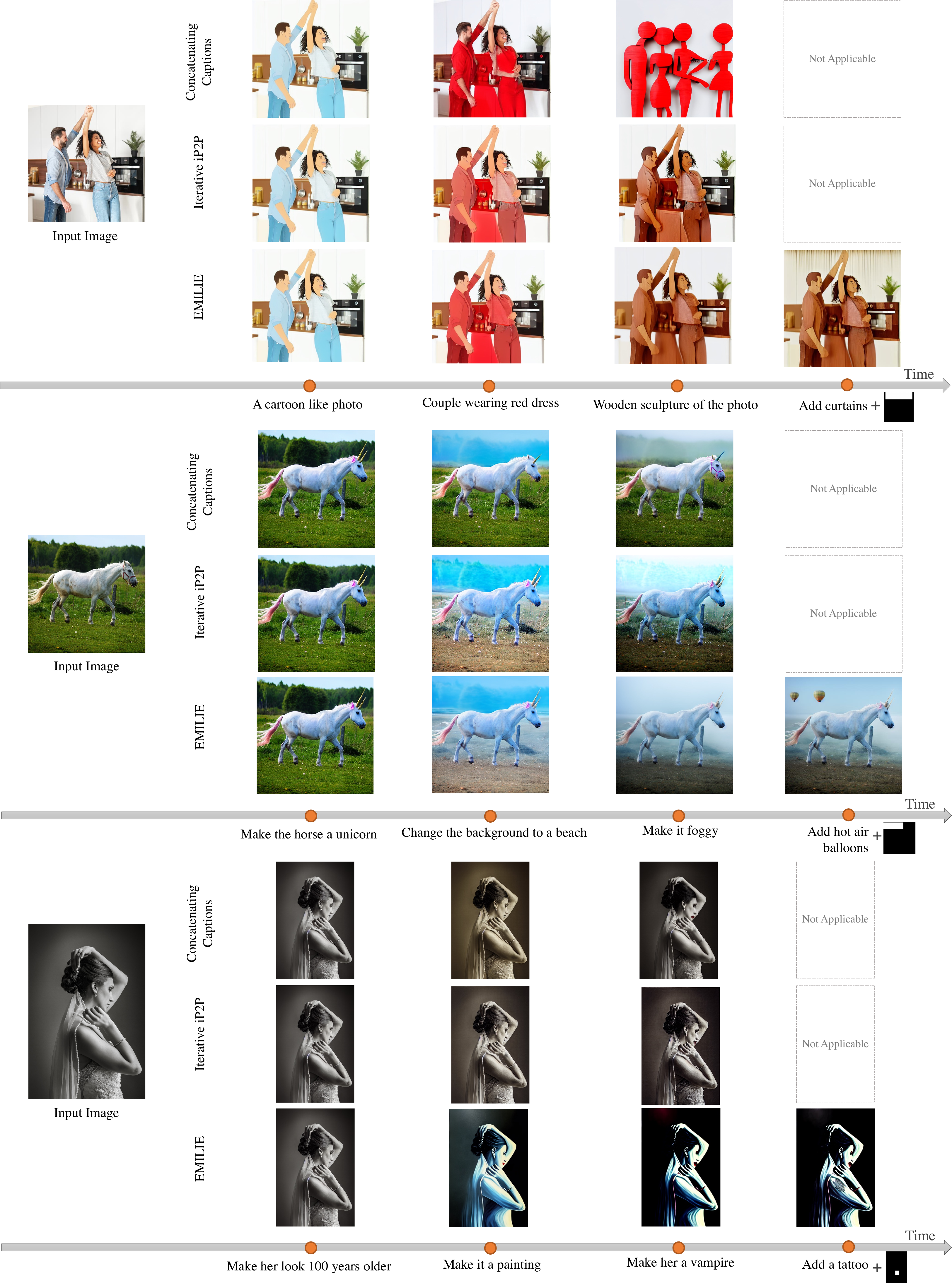}
  \caption{We showcase more qualitative comparisons with baselines. We see that \ours is able to produce more realistic and meaningful edits when compared to concatenating all the instructions seen so far, and iteratively using Instruct Pix2Pix \cite{brooks2022instructpix2pix}.
  }
  \label{fig:suppl_results_2}
  \vspace{-10pt}
\end{figure*}

\cref{fig:suppl_results_1,fig:suppl_results_2,fig:suppl_results_3} showcases more qualitative evaluation of \ours. We compare with two top performing baselines: 1) concatenating all the edit instructions that we have seen so far and applying them to the original image, and 2) iteratively passing the edited image through Instruct Pix2Pix \cite{brooks2022instructpix2pix} to be updated by the newer edit instruction. We see that \ours is able to create images with lower noise levels and is more visually appealing and semantically consistent. 

\section{Failure Cases}
\begin{figure}[b]
\vspace{-10pt}
  \centering
\includegraphics[width=1\columnwidth]{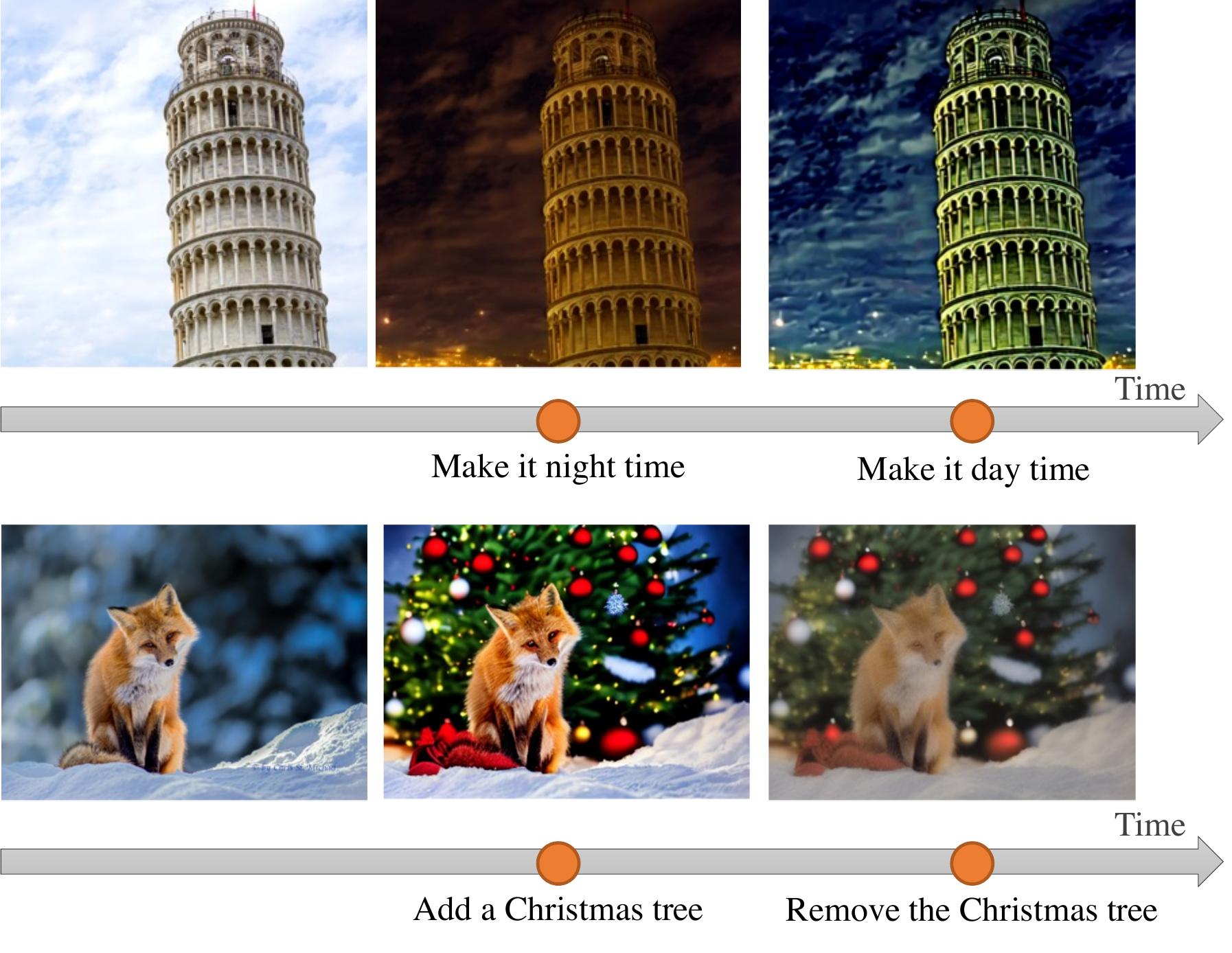}
\vspace{-20pt}
  \caption{\ours is not able to do identity preservation across edits that reverse concepts and conflict each other. 
  }
  \label{fig:failurecasesjkj}
  \vspace{-5pt}
\end{figure}

\begin{figure*}
  \centering
\includegraphics[width=1.85\columnwidth]{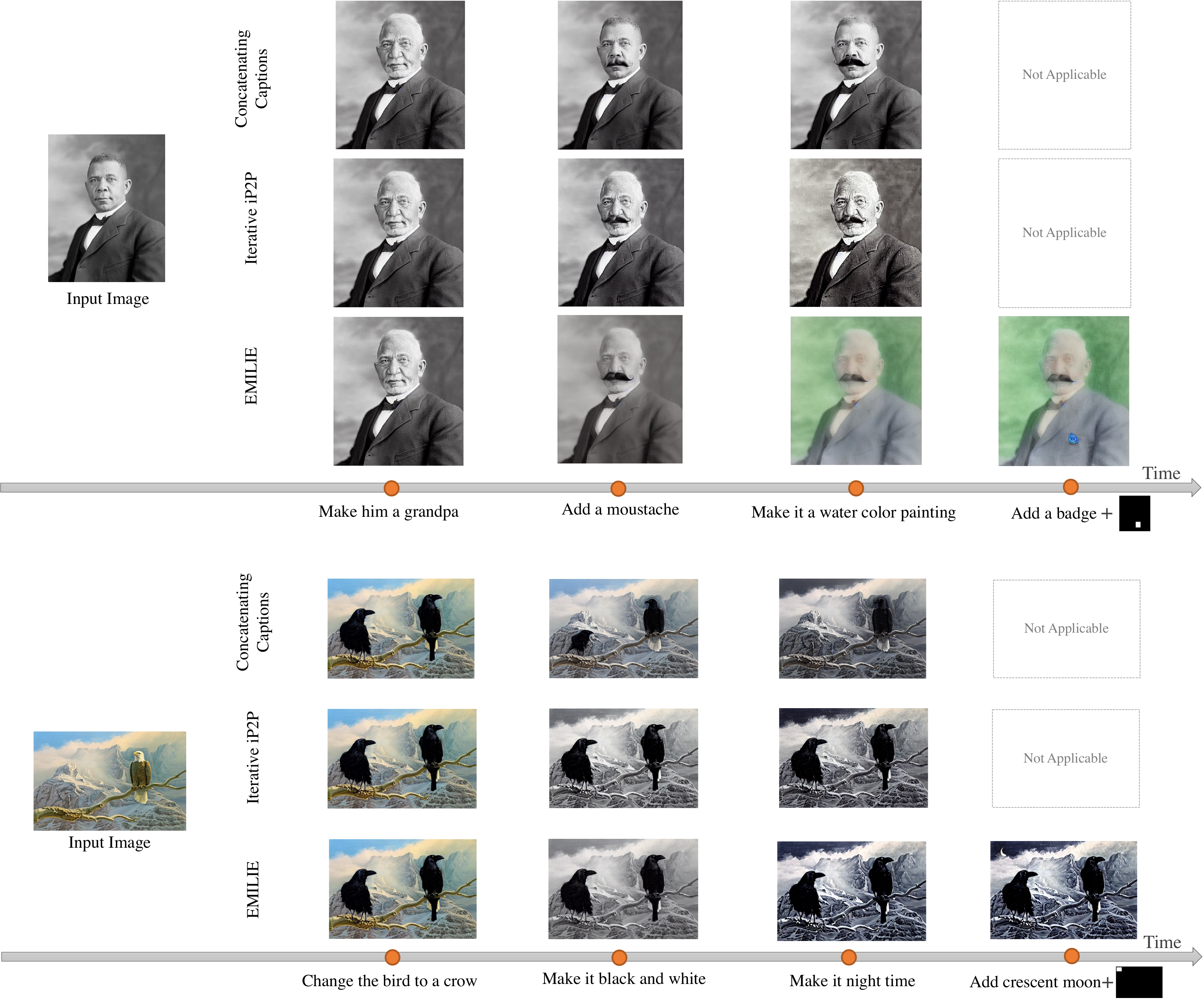}
  \caption{We showcase more qualitative comparisons with baselines. We see that \ours is able to produce more realistic and meaningful edits when compared to concatenating all the instructions seen so far, and iteratively using Instruct Pix2Pix \cite{brooks2022instructpix2pix}.
  }
  \label{fig:suppl_results_3}
\end{figure*}

\ours has shortcomings too. \cref{fig:failurecasesjkj} shows one such setting where \ours fails to undo edit instructions. It is natural for the edit instructions to be conflicting to each other. For example, we might add a Christmas tree at step 1, and later plan to remove it. This is hard for the method to do. 
Also, in some cases, the model makes inconsistent changes. For instance, in \cref{fig:suppl_results_3}, the mustache that was added in step 2 should have had ideally been gray. Though the edit instruction was to change the existing bird to the crow, the model indeed added a new bird too.
This is largely an artifact of the base editing framework \cite{brooks2022instructpix2pix} that we build on. These are interesting future research directions.


\end{document}